\documentclass[a4paper,article,10pt]{memoir}
\usepackage[english]{babel}
\usepackage{CCLAuthor}
\usepackage[reqno]{amsmath}
\usepackage{amssymb}
\usepackage{amsthm}  
\theoremstyle{plain}

\theoremstyle{definition}

\numberwithin{theorem}{chapter}
\usepackage[round,authoryear]{natbib} 
\usepackage[]{graphicx}               
\begin{document}
%
%
%
\title{Cuspidal Robots}
%
%
\author{%
    Philippe Wenger 
    \\ \smallskip\small
  
    IRCCyN, CNRS, Nantes, France                   
 
    }
    \maketitle
%
%
%
%
    \begin{abstract}
    This chapter is dedicated to the so-called cuspidal robots, i.e. those robots that can move from one inverse geometric solution to another without meeting a singular confuguration. This feature was discovered quite recently and has then been fascinating a lot of researchers. After a brief history of cuspidal robots, the chapter provides the main features of cuspidal robots: explanation of the non-singular change of posture, uniqueness domains, regions of feasible paths, identification and classification of cuspidal robots. The chapter focuses on 3-R orthogonal serial robots. The case of 6-dof robots and parallel robots is discussed in the end of this chapter.
		
    \end{abstract}

\CCLsection{Introduction}
		
\begin{figure}[htbp]
            \centering
        \includegraphics[width=0.5\linewidth]{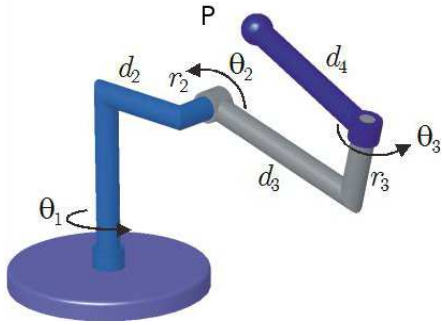}
        \caption{A serial robot with three revolute joints.}
	\label{fig1}
    \end{figure}

		\CCLsubsection{Preliminaries}
		This chapter deals with the so-called cuspidal robots. It is focused on serial, open-chain robots. 
For these robots, the geometric model can be defined by an input map f, such that $\textbf{X}=f(\textbf{q})$, where q is the vector of joint variables (the inputs), which are all actuated. Vector \textbf{X} contains the coordinates of the end-effector pose (the outputs). The robots are assumed non-redundant, namely, the number of joint variables ($n$) is equal to the number of coordinates ($m$) that describe the end-effector pose $(n=m)$. 
The configuration space of these robots can be restricted to their joint space, which is fully defined by the actuated joints. As a consequence, the only singularities are defined by the input singularities. Unless specified, the word singularity will stand for input singularity in this chapter. 
Figure~\ref{fig1} shows a 3-R robot (made of three revolute joints). The input variables are the three joint coordinates $\theta_1$, $\theta_2$ and $\theta_3$. The output variables are defined by the three Cartesian coordinates $x$, $y$ and $z$ of a reference point $P$ fixed on the last link. For this robot, $n=m=3$ and its joint space is a 3-Torus if the joints are unlimited (a 3-D box if they are limited). Generically, the singularities form a set of surfaces in the joint space. These surfaces divide the joint space into a set of singularity-free connected regions called aspects, as defined by \citet{Borrel86}.

		As it will be shown further, a cuspidal robot is a robot that can move from one of its inverse geometric solution to another without encountering a singularity. This feature was discovered quite recently and has then been fascinating a lot of researchers. Before this discovery, it was thought that any robot should necessarily cross a singularity during a change of solution, like, e.g., when an anthropomorphic robot (a Puma type robot) switches from the elbow up to the elbow down configuration by passing through the fully outstretched (or folded) singular configuration (see section \ref{sec:def_posture}). A brief history of cuspidal robots is presented next.
		
		\CCLsubsection{Brief history of cuspidal robots}
		In 1986, a mathematical proof was provided in order to confirm the well-established belief that any non-redundant serial robot should cross a singularity when moving from one solution to another \citep{Borrel86}. This proof was based on the implicit function theorem and some topological arguments. The proof was in fact incorrect but at that time nobody noticed it as it was just confirming a fact that everybody had admitted before. 
The very first mention of a singularity-free change of solution dates back to 1988, in a talk by V. Parenti-Castelli and C. Innocenti (University of Bologna, Italy) presented during the first conference on Advances in Robot Kinematics that took place in Ljubljana \citep{ParentiARK88}. For two different types of 6-R (revolute-jointed) robots, the authors were able to build a singularity-free path joining two inverse solutions. 
It was such a big challenging of a well-established fact that nobody trusted them at that time. 
In a separated work conducted in Stanford University, J. W. Burdick showed in its PhD thesis several 3-R robots also able to move from one solution to another without encountering a singularity \citep{BurdickPhD}. These results were not published outside the PhD thesis. Nothing appeared during the next four years before this behavior was then formalized in 1992 by \citet{Wenger92}. It was clearly shown in this work why even a simple 3-R robot can move from one of its solution to another without meeting a singularity. It was not easy to understand why the proof given in \citep{Borrel86} was incorrect as it seemed mathematically sound. The main problem was that the proof resorted to a hypothesis that turned out to be not true in all robots. The detailed explanation is a bit more tricky and is not reported there. 
The belief that a change of solution should necessarily be singular was so strongly rooted in people\text{'}s mind that it took several years before the research community started to accept the existence of cuspidal robots. The word cuspidal was defined after 1995 when it was shown that the existence of a cusp point in the singularity locus of the robot indicates that this robot can change its posture without encountering a singularity by encircling this cusp point \citep{ElOmri95}, see sections \ref{sec:non-sing_sing} and \ref{sec:identif}.

		\CCLsubsection{Questions of interest}
		\label{sec:quest}
When the research community started to admit the truth in the second part of the 90’s, it was faced with a lot of questions, such as:

\begin{itemize}[--]
\item How can a non-singular change of solution be accomplished for a given robot? 
\item What are the consequences on path-planning? 
\item Is this ability possible only for specific robots or is it more general? 
\item Given a new robot, how to know if it will have this ability? 
\item Is it possible to set some conditions for a robot to have this ability? 
\item Is it possible to enumerate all the cuspidal robots? 
\end{itemize}

The first two issues are very important for control, while the others are very important for the designer. As it will be shown further, a change in geometric parameters of a non-cuspidal robot may render it cuspidal. Interestingly, this fact can be illustrated with the story of the IRB 6300C robot launched by ABB in 1996. This new manipulator was specially designed for the car industry to minimize the swept volume. The only difference from the Puma was the permutation of the first two link axes, resulting in a manipulator with all its joint axes orthogonal. Commercialization of the IRB 6400C was stopped one year later. Informal interviews with robot customers at that time revealed difficulties in planning offline trajectories using Robotic-CAD systems for this robot. In fact it was shown later that the IRB 6400C robot  turns out to be cuspidal.

\CCLsubsection{Chapter outline}
In the rest of this chapter, the singular and non-singular changes of posture are analyzed in details. Next the feasibility of trajectories in cuspidal robots are investigated and the maximal regions of feasible paths in the workspace are defined. The identification, enumeration and classification of 3-dof (degree-of-freedom) cuspidal robots and non-cuspidal robots are then addressed. Eventually, the case of 6-dof robots and parallel robots are discussed.

\CCLsection{Postures}
		
\begin{figure}[htbp]
            \centering
        \includegraphics[width=0.5\linewidth]{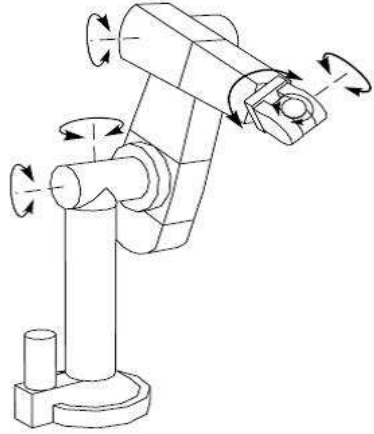}
        \caption{An anthropomorphic robot.}
	\label{fig2}
    \end{figure}

\begin{figure}[htbp]
            \centering
        \includegraphics[width=1\linewidth]{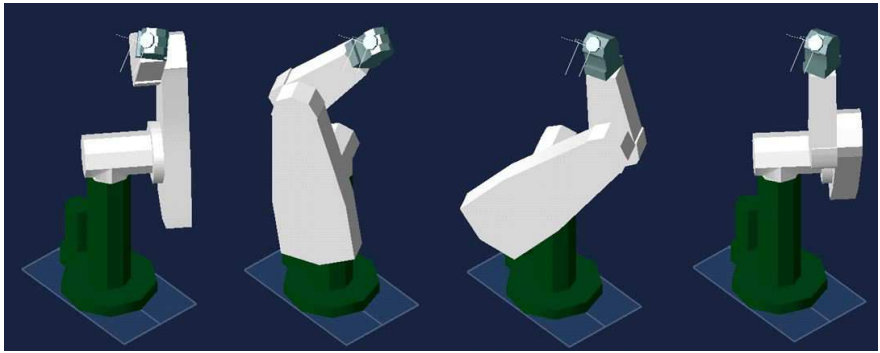}
        \caption{Four postures for the anthropomorphic robot, from left to right: (elbow up, shoulder left), (elbow up, shoulder right), (elbow down, shoulder right), (elbow down, shoulder left).}
	\label{fig3}
    \end{figure}

		\CCLsubsection{Definition}
		\label{sec:def_posture}
		A posture is associated with one solution to the inverse geometric problem of a robot. This word refers to the way a given robot places its links in space to reach a given frame. For usual industrial robots, a posture can be easily identified geometrically. For example, it is well known that an anthropomorphic robot shown in Figure~\ref{fig2}, has at most eight distinct inverse geometric solutions, which are associated with eight postures. These postures can be identified geometrically by the configuration of the elbow (up or down), the shoulder (right or left) and the wrist (flip or no flip). The total number of combinations amount to $2^3=8$ distinct postures. Figure~\ref{fig3} shows four of the eight postures obtained in the flip configuration of the wrist. It is well known that the only way for this robot to switch from any of its posture to another is to cross a singular configuration.
		
\begin{figure}[htbp]
            \centering
        \includegraphics[width=0.5\linewidth]{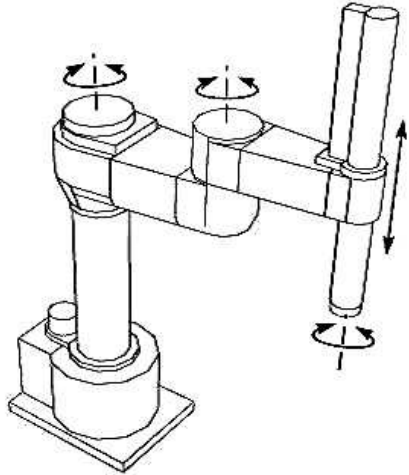}
        \caption{A SCARA robot.}
	\label{fig4}
	    \end{figure}

\begin{figure}[htbp]
            \centering
        \includegraphics[width=0.9\linewidth]{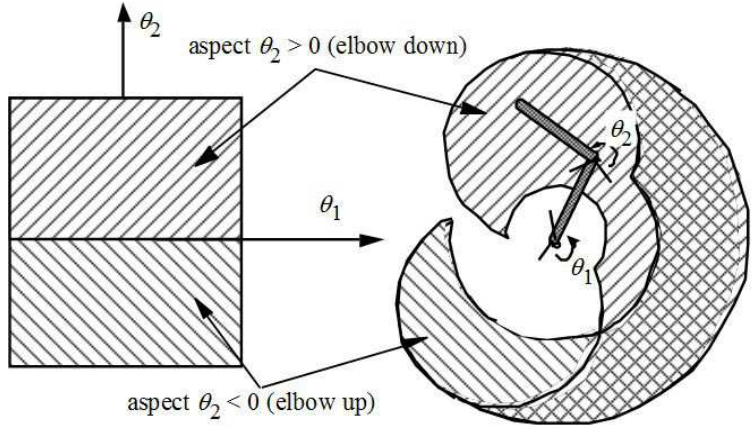}
        \caption{Aspects of the SCARA robot in the joint space (left) and workspace (right).}
	\label{fig5}
	    \end{figure}

		\CCLsubsection{Postures and aspects – case 1 : non-cuspidal robots}
		\label{sec:post_asp_NC}
		The aspects are the largest singularity-free connected regions of the joint space \citep{Borrel86}. They are bounded by the singularity surfaces (or hyper-surfaces for a robot with more than 3 joints) and by the joint limits when they exist. Additional boundaries may occur in the presence of obstacles \citep{ElOmri96}. Before cuspidal robots were discovered, one posture was thought to be uniquely associated with one aspect. Indeed, since the robot was thought to cross necessarily a singularity when moving from one posture to another, there could not be more than one posture in each of its aspects. This is true for most usual industrial robots like the anthropomorphic robot and the SCARA robot shown in Figure~\ref{fig3} and Figure~\ref{fig4}, respectively. The SCARA robot is a well-known industrial robot (Figure~\ref{fig4}). This 4-dof robot can produce three translations in space and one rotation about a vertical axis. This motion is known as the Sch\"onflies motion, first studied by the German mathematician Arthur-Moritz Sch\"onflies (1853--1928). This motion is suitable for pick-and-place tasks.
	The determinant of the Jacobian matrix $det(J)$ vanishes whenever $\sin(\theta_2)=0$, where $\theta_2$ is the second revolute joint variable. Accordingly, a singularity occurs whenever $\theta_2=0$ or $\theta_2=\pi$, namely, when the arm is fully extended or fully folded. This robot has two aspects defined by $\theta_2>0$ and $\theta_2<0$, respectively. Moreover, the inverse geometric model of this robot admits two solutions, associated with the two postures “elbow up” and “elbow down”, respectively. Fig. 5, left, shows the two aspects in $(\theta_1, \theta_2)$. The other joint variables can be ignored since they do not play any role in the singularities and aspects. The robot is assumed to have joint limits and its joint space is thus a square in $(\theta_1, \theta_2)$. Figure~\ref{fig5}, right, shows the aspects projected in the workspace and the robot depicted in an elbow down posture. There is only one inverse geometric solution in each aspect; namely, a posture is associated with one aspect.
	
\begin{figure}[htbp]
            \centering
        \includegraphics[width=1\linewidth]{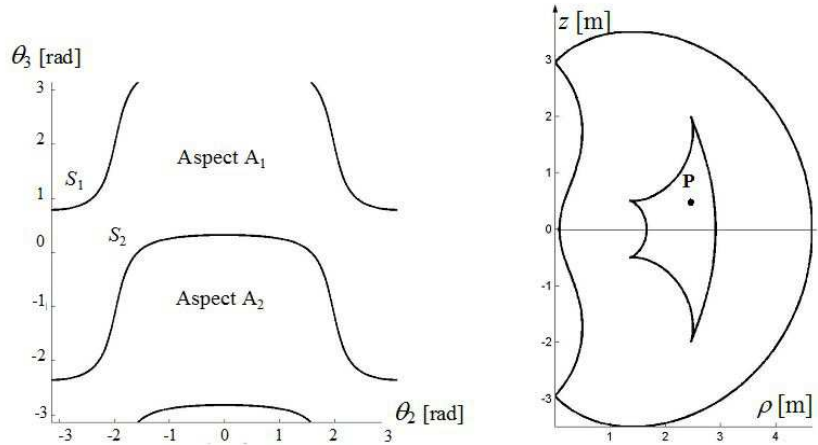}
        \caption{Singularity curves for the 3-R cuspidal robot shown in fig. 1 (left) and its projection in the workspace (right).}
	\label{fig6}
	    \end{figure}

\begin{figure}[htbp]
            \centering
        \includegraphics[width=0.7\linewidth]{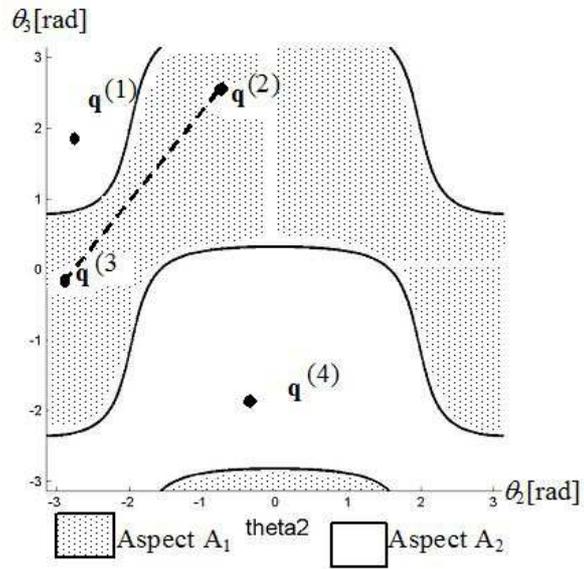}
        \caption{Two solutions in each aspect and a non-singular path joining two solution in $A_1$.}
	\label{fig7}
	    \end{figure}

\begin{figure}[htbp]
            \centering
        \includegraphics[width=0.9\linewidth]{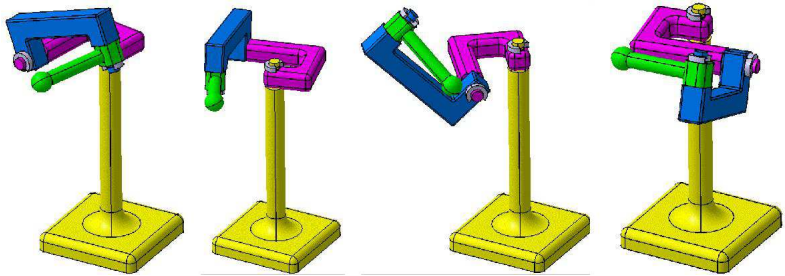}
        \caption{The four postures of the 3-R cuspidal robot at point $P$.}
	\label{fig8}
	    \end{figure}

		\CCLsubsection{Postures and aspects – case 1 : cuspidal robots}
		\label{sec:post_asp_C}
A cuspidal robot has more than one inverse geometric solution in at least one of its aspects. In other words, it can move from one posture to another without meeting a singularity. The 3-R robot with mutually orthogonal joint axes shown in fig. 1 turns out to be cuspidal because this robot has two aspects and four inverse geometric solutions, two in each aspect. The determinant of the Jacobian matrix $\textbf{J}$ of this robot can be written as:

\begin{equation}
det ({\textbf{J}}) = ({d_3} + {c_3}{d_4})({c_2}({s_3}{d_3} - {c_3}{r_2}) + {s_3}{d_2}) \label{eq:1}
\end{equation}		
		
		where $c_i=\cos(\theta_i)$ and  $s_i=\sin(\theta_i)$, $i=2,3$. The geometric parameters $d_2$, $d_3$, $d_4$ and $r_2$, shown in Figure~\ref{fig1}, are the Modified Denavit-Hartenberg as defined in \citep{Symoro}. It can be noticed that the first joint variable $\theta_1$ does not appear in this determinant. This is always the case in any serial robot whose first joint is revolute since in this case the singularities do not depend on $\theta_1$. The singularity surfaces of our robot, which are the zero sets of $det(\textbf{J})=0$, can then be plotted in $(\theta_1, \theta_2)$ where they appear as curves. The first factor may vanish only when $d_3<d_4$, namely when the last link is longer than the second link. If it is not the case, the singularity curves are defined by the zeros of the second factor only. Figure~\ref{fig6} (left) shows the singularity curves obtained when $d_2=1$, $d_3=2$, $d_4=1.5$ and $r_2=1$ and for unlimited joints.
			Note that since the robot is assumed to have no joint limits, its joint space has the structure of a Torus and the singularity curves are defined on the Torus $T(\theta_2,\theta_3)$. For more simplicity, a plane picture of the joint space is shown in fig. 6 and the reader may retrieve the topology of $T(\theta_2,\theta_3)$ by identifying the opposite sides of the square. Having this important notice in mind, it is easy to see that the singularities form two closed curves $S_1$ and $S_2$ and define only two aspects. The singularities can be also plotted in the workspace. For our 3-R robot, the workspace is of dimension 3 but it is sufficient to plot a planar cross section passing through the first joint axis because we know that the singularities do no depend on $\theta_1$. The workspace can then be plotted in the plane $(\rho, z)$, where 
$\rho  = \sqrt {{x^2} + {y^2}}$  (Figure~\ref{fig6}, right). The outside boundary of the workspace $WS_2$ is the image of $S_2$ while the internal boundary $WS_1$ is the image of $S_1$. The internal boundary is made of four lines that merge at four cusp points. This internal boundary separates two regions: the inner region and the outer region. It can be verified that the robot has four inverse geometric solutions in the inner region but it has only two solutions in the outer region. We solve the inverse geometric model at a point $P(\rho=2.5, z=0.5)$ in the inner region. The four solutions are (in radians): $q^{(1)}=[-1.8, -2.8, 1.9]^t, q^{(2)}=[-0.9, -0.7, 2.5]^t, q^{(3)}=[-2.9, -3, -0.2]^t$ and $q^{(4)}=[0.2, –0.3, –1.9]^t$. It is apparent from fig. 7 that $q^{(2)}$ and $q^{(3)}$ lie in the same aspect $A_1$. It is then possible to link $q^{(2)}$ and $q^{(3)}$ by a non-singular path as shown in Figure~\ref{fig7}. Keeping in mind that the opposite sides of the square should be identified, one can see that $q^{(1)}$ and $q^{(4)}$ also lie in one single aspect ($A_2$).
Contrary to their non-cuspidal counterparts, the anthromorphic robot and the SCARA robot, the postures of this 3-R cuspidal robot are difficult to identify geometrically, as shown in Figure~\ref{fig8}.
		
\begin{figure}[htbp]
            \centering
        \includegraphics[width=0.9\linewidth]{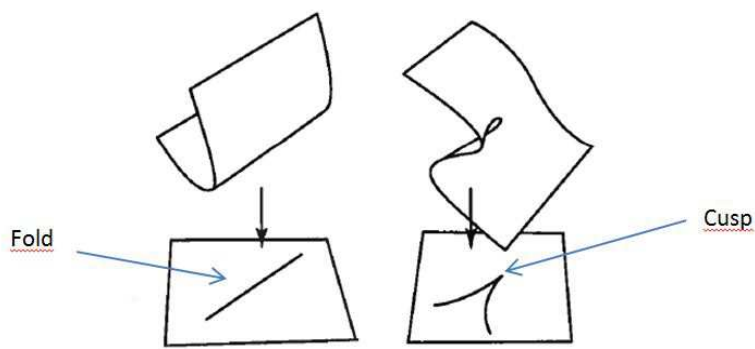}
        \caption{A fold (left) and a cusp (right). Adapted from \citep{Gibson}.}
	\label{fig9}
	    \end{figure}

\begin{figure}[htbp]
            \centering
        \includegraphics[width=0.5\linewidth]{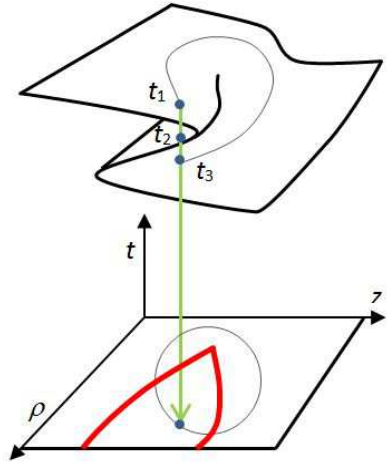}
        \caption{A non-singular path joining two distinct solutions.}
	\label{fig10}
	    \end{figure}

\begin{figure}[htbp]
            \centering
        \includegraphics[width=0.5\linewidth]{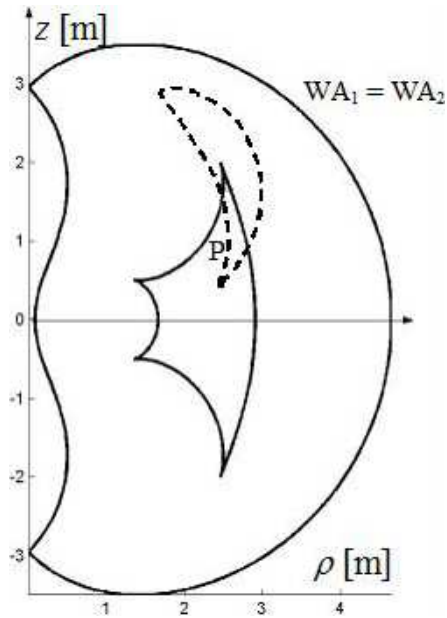}
        \caption{Non-singular posture changing path in the workspace.}
	\label{fig11}
	    \end{figure}

\begin{figure}[htbp]
            \centering
        \includegraphics[width=0.9\linewidth]{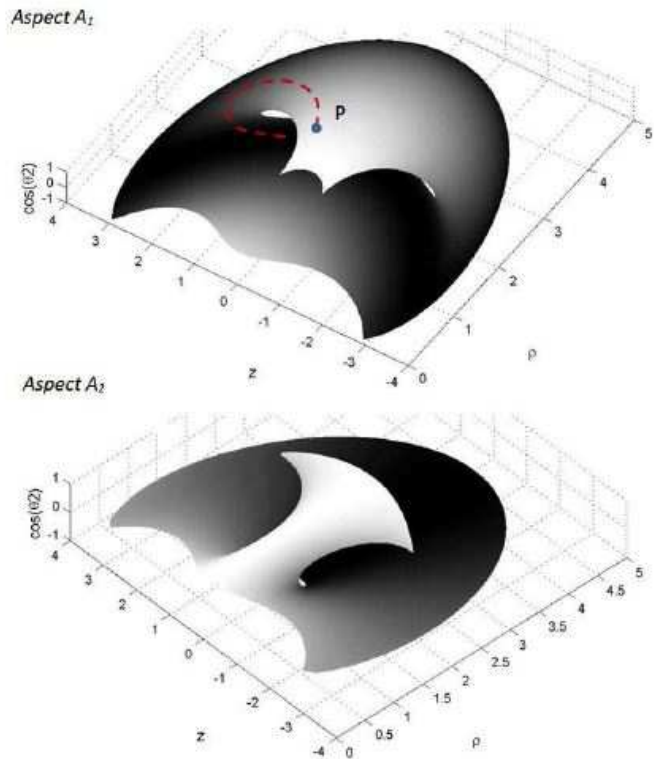}
        \caption{The splitted configuration space cut along its folds and shown in the two aspects. A non-singular posture changing path is shown in aspect A1 (above).}
	\label{fig12}
	    \end{figure}

		\CCLsubsection{Non-singular change of posture vs singular change of posture}
		\label{sec:non-sing_sing}
The question of how a cuspidal robot can move from one of its postures to another without encountering a singularity is of primary interest and has puzzled the research community for several years. People thought that the robot should meet a special transitory state between the two postures during a change of posture. If it is not a singularity, what kind of transitory state should it be? In fact no special transitory state is met during a non-singular change of posture and this change is accomplished quite smoothly. To better understand this behavior, it might be helpful to refer to the mappings from a surface onto a plane as formalized by Whitney in 1955. Whitney showed that only two types of stable singularities may occur when a surface is projected onto a plane: the fold and the cusp \citep{Whitney}. A fold arises when the surface is simply folded like in Figure~\ref{fig9}, left. Its projection onto a plane defines a simple line. A cusp arises when the surface is folded twice as in Figure~\ref{fig9}, right. The projection onto a plane gives rise to two fold lines that merge at a cusp point.
Now, the non-singular change of posture can be interpreted by considering that the doubly-folded surface shown on the right of Figure~\ref{fig9} is the configuration space $F(\rho, z, t)=0$ of a cuspidal robot, where $\rho$ and $z$ are the output variables and $t$ is associated to one of the joint variables. The projection onto the plane corresponds to the workspace of the robot. The two red lines merge at a cusp point and define a region where the robot admits three solutions $t_1$, $t_2$ and $t_3$, each one being associated with on one of the three layers of the doubly-folded configuration space surface. Figure~\ref{fig10} shows a path connecting two solutions $t_1$ and $t_3$ (from layer 1 to layer 3) at a given point $(\rho, z)$ of this 3-solution region. This path goes “smoothly” between the solutions and no special transitory configuration is met. Moreover, it turns out that the path goes around the cusp, a fact that will prove of primary importance for the characterization of cuspidal robots. One should be aware that if the projection of the path onto the plane does cross the two red line segments, the path itself never meets the corresponding folds of the configuration space and it is really singularity-free. 
			Coming back to our cuspidal 3-R robot, the non-singular posture changing path shown in Figure~\ref{fig7} is now projected in the workspace cross section (Figure~\ref{fig11}).
		Like in Figure~\ref{fig10}, the path encircles a cusp point. The same remark as above also holds here: no singularity is met even if two segments of the inner boundary are crossed. The shape of the configuration space of the robot is a bit more complicated than the one illustrated in Figure~\ref{fig10}. It is easier to cut it along its folds and to show separately its restrictions to the aspects. Figure~\ref{fig12} shows the configuration space restricted to the aspect $A_1$ (above) and aspect $A_2$ (below). To get the complete configuration space, these two parts should be glued together along their common edges. In fact the complete configuration space folds along these edges. The fact that there are four inverse geometric solutions in the inner region and only two in the outer region is in accordance with the number of “layers” of the configuration space in these regions. The non-singular posture changing path of Figure~\ref{fig11} is shown in red dashed lines. This path goes smoothly from one layer to another like in Figure~\ref{fig10} by going around a cusp point. Note that in $P$ there are two more solutions in aspect $A_2$ and a non-singular posture changing path between these two solutions will have to encompass one of the two opposite cusp points.
A singular posture changing path can also be defined in $P$. In the configuration space, such a path should go from one layer of an aspect to a layer of another aspect by moving on a fold. The “splitted” model of the configuration space shown in Figure~\ref{fig12} does not make it possible to show this path on the same image. Starting from P in aspect $A_1$, the path will have to go towards one of the edges. Once on the edge, the path is in fact on the fold that enables it to enter the second aspect $A_2$ and a singularity is encountered, where two inverse geometric solutions merge. Physically, the robot behaves like if it was “bouncing back” against a workspace boundary (think of the SCARA robot that has to bounce back in the fully outstretched configuration when it moves from the elbow up to the elbow down posture).

\begin{figure}[htbp]
            \centering
        \includegraphics[width=0.5\linewidth]{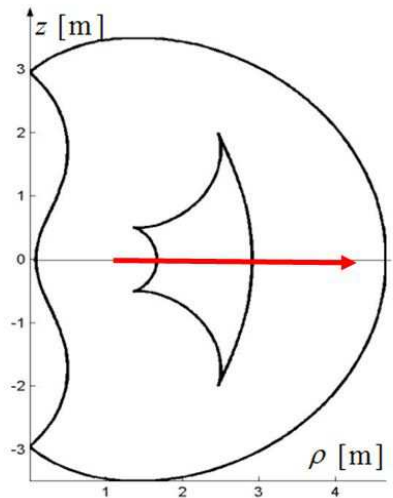}
        \caption{An infeasible path for the cuspidal robot.}
	\label{fig13}
	    \end{figure}

\CCLsection{Path feasibility}			
		\CCLsubsection{Showing an example}
		\label{sec:ex}
		 Another important feature of cuspidal robots is their inability to track certain prescribed paths in their workspace, even in the absence of joint limits or obstacles. Let us consider the straight line path shown in red in Figure~\ref{fig13}. The arrow indicates the direction of motion. The robot is unable to track this path. The reason is that the robot will have to stop on the second crossed boundary line. Indeed, Figure~\ref{fig12} shows that an edge will be always met, should the robot start from aspect $A_1$ or $A_2$. Since when arriving on an edge the robot has no other choice than bouncing back, it cannot track the path further. Note that the path is also infeasible in the reverse direction.
		In usual non-cuspidal robots met in industrial robotic sites, path infeasibility is usually due to a physical obstruction such as a joint limit or a collision. The above example shows that a path can be infeasible even if the robot has not joint limits and if there are no collisions. 
The following section proposes a way of defining the regions of feasible paths in the workspace for cuspidal robots.
			
\begin{figure}[htbp]
            \centering
        \includegraphics[width=0.8\linewidth]{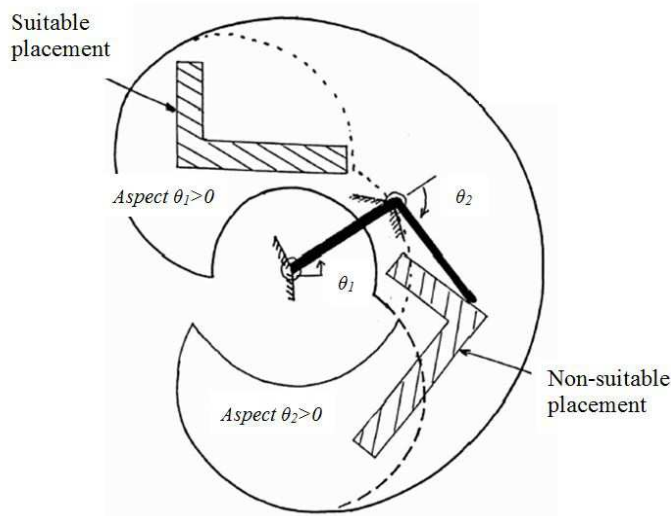}
        \caption{Use of the aspects for cutting trajectories for a non-cuspidal robot.}
	\label{fig14}
	    \end{figure}

		\CCLsubsection{Regions of feasible paths for non-cuspidal robots}
		\label{sec:path}
		The question of whether a prescribed path in the workspace can be fully tracked or not is of primary interest for many robotic applications such as process tasks. Think of an arc-welding task in which the soldering torch attached to the robot end-effector must realize a continuous welding seam. The welding seam must be fully realized in one step at a constant velocity. This is the reason why P. Borrel introduced the notion of aspects (see section \ref{sec:post_asp_NC}) when he started to work of the design of a Robotic CAD interface with CATIA for robots offline programming \citep{Borrel86} in 1986. For a non-cuspidal robot, there is only one posture in each aspect. This means that the kinematic map is one-to-one from each aspect onto the workspace. In other words, the aspects are the largest uniqueness domains of the joint space for a non-cuspidal robot. This property is useful to analyze the feasibility of paths. Indeed, since the kinematic map is one-to-one on each aspect, the pre-image of any connected set lying in the image of an aspect in the workspace defines a connected set in this aspect. This means that any continuous path lying in the image of an aspect is the image of a continuous path in the joint space. In other words, any path in the image of an aspect is feasible by the robot. This means that the images of the aspects are the (maximal) regions of feasible paths in the workspace for a non-cuspidal robot. In a robotic CAD system, it is this very convenient to determine the regions of feasible paths in the workspace since the user will then know where to define the prescribed paths during the off-line programming of a robot. Referring to Figure~\ref{fig5}, this SCARA robot has two regions of feasible paths in the workspace that overlap. Suppose that this SCARA robot is used to cut L-shape plates. The robot tool must be able to follow the contour of the shape continuously. Figure~\ref{fig14} shows the two regions of feasible paths along with a suitable and a non-suitable placement of the plates in the workspace. The one that is fully included in the image of an aspect (above) is suitable while the other (below) is not.

\begin{figure}[htbp]
            \centering
        \includegraphics[width=1\linewidth]{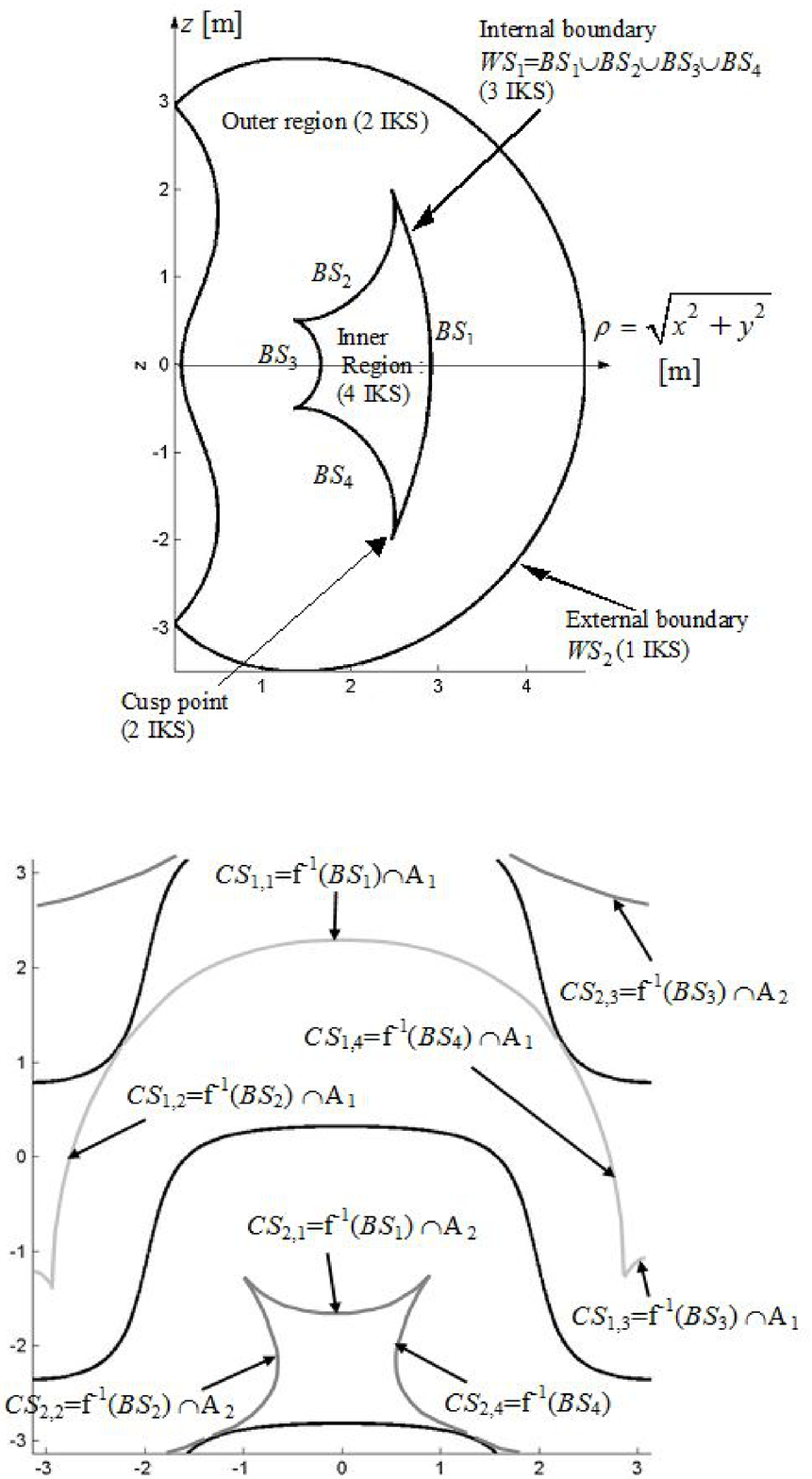}
        \caption{The characteristic surfaces (below) of the cuspidal robot of fig. 1 and correspondence with the segments of the internal boundary in the workspace (above).}
	\label{fig15}
	    \end{figure}

			For a cuspidal robot, the aspects do not define uniqueness domains any more. Accordingly, the regions of feasible paths are not the images of the aspects in the workspace. New uniqueness domains must be defined. 
		\CCLsubsection{Characteristic surfaces}
		\label{sec:char}
Since the singular surfaces in the joint space do not separate all the inverse geometric solutions, new separating surfaces should exist. The set obtained by calculating the non-singular inverse geometric solutions for all points on an internal boundary forms a set of non-singular surfaces in each aspect. These surfaces are called the characteristic surfaces \citep{Wenger92}. A set of characteristic surfaces is associated with one aspect and they separate the inverse geometric solutions in each aspect, as shown further. A general definition of the characteristic surfaces can be set as follows, which stands for any non-redundant manipulator. Let $A_i^*$  be the boundary of aspect $A_i$. The characteristic surfaces 
\{$C{S_i}$\} associated with $A_i$ are:
\begin{equation}
 \{C{S_i}\} = f^{-1}(f(A_i^*)) \cap A_i \label{eq:2}
\end{equation}
			where $f(A_i^*)$ is the image of $A_i^*$ under the forward kinematic map and $f^{-1}(f(A_i^*))=\{\textbf{q} / f(\textbf{q})\in f(A_i^*)\}$. Note that since an aspect is defined as an open set, $A_i$ does not contain its boundary i.e. $A_i^*\cap A_i=\varnothing$ thus \{$CS_i$\} might be empty (note that if this is the case for all its aspects, the robot is not cuspidal). Since the general definition of the characteristic surfaces is not algebraic in nature, it is difficult to derive an algebraic expression of \{$CS_i$\}  that would be easy to handle. It should be thus calculated numerically.
For the cuspidal robot of Figure~\ref{fig1} studied in Figures~\ref{fig6}--\ref{fig8}, the set of characteristic surfaces \{$CS_i$\} associated with $A_i^*$ can be defined as $\{C{S_i}\} = f^{-1}(WS_i) \cap A_i$, where $f^{-1}(WS_i)=\{\textbf{q} / f(\textbf{q})\in WS_i\}$. Since $WS_1=BS_1 \cap BS_2 \cap BS_3 \cap BS_4$ (Figure~\ref{fig15} above), we can write $f^{-1}(WS_1)=f^{-1}(BS_1) \cup f^{-1}(BS_2) \cup f^{-1}(BS_3) \cup f^{-1}(BS_4)$.
Each set $f^{-1}(BS_j)$ has two components $CS_{1,j}=f^{-1}(BS_j)\cap A_1$ and $CS_{2,j}=f^{-1}(BS_j)\cap A_2$ in $A_1$ and $A_2$, respectively. Thus, the two sets of characteristic surfaces can be written as $\{C{S_1}\}=CS_{1,1} \cup CS_{1,2} \cup CS_{1,3} \cup CS_{1,4}$ and $\{C{S_2}\}=CS_{2,1} \cup CS_{2,2} \cup CS_{2,3} \cup CS_{2,4}$ respectively (Figure \ref{fig15} below). Any two adjacent segments of a set of characteristic surfaces meet at a preimage of a cusp point. This preimage is either a singular configuration or a non-singular one. When the preimage is singular, the segments meet tangentially to a singular surface, whereas a non-singular preimage of a cusp point forms a cusp in an aspect. For example, the segments $CS_{1,1}$ and $CS_{1,2}$ meet tangentially to the singularity surface $S_1$ at the singular preimage of the cusp point that links $BS_1$ and $BS_2$ in the workspace. The remaining non-singular preimage of this cusp point connects the segments $CS_{2,1}$ and $CS_{2,2}$ and forms a cusp in aspect $A_2$ (Figure~\ref{fig15} below).		
		
\begin{figure}[htbp]
            \centering
        \includegraphics[width=0.9\linewidth]{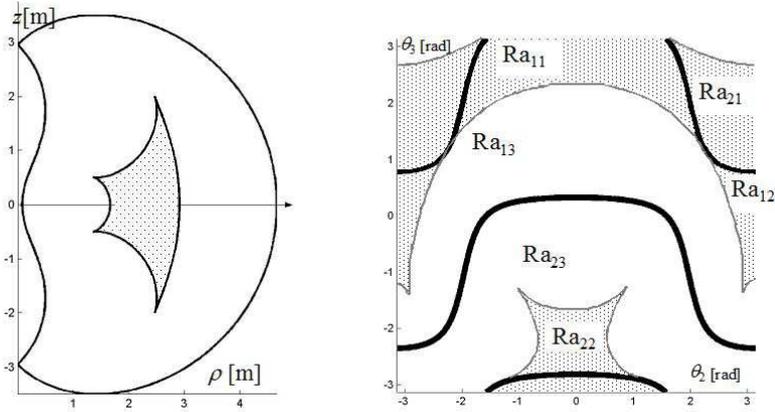}
        \caption{The characteristic surfaces divide the aspects into smaller sets.}
	\label{fig16}
	    \end{figure}

\begin{figure}[htbp]
            \centering
        \includegraphics[width=0.8\linewidth]{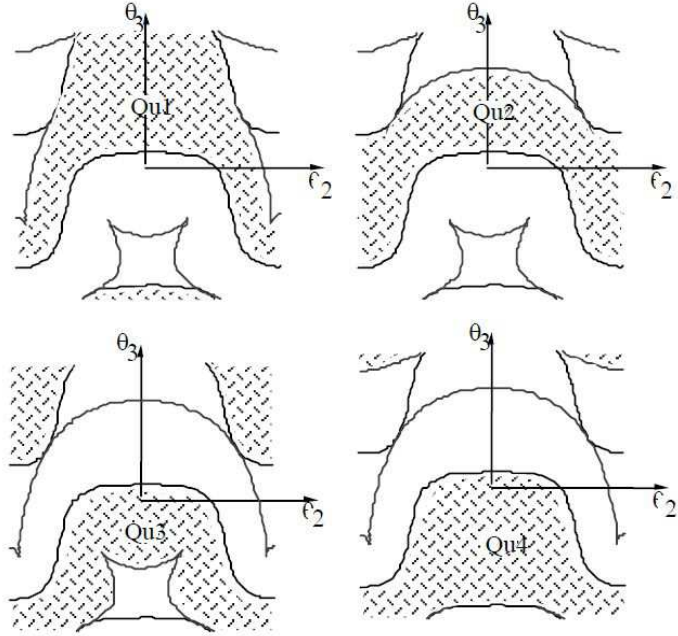}
        \caption{The four uniqueness domains for the robot of Figure~\ref{fig1}.}
	\label{fig17}
	    \end{figure}

		\CCLsubsection{Uniqueness domains and regions of feasible paths for cuspidal robots}
		\label{sec:unique}
The characteristic surfaces induce a partition of each aspect into smaller sets $Ra_{ij}$. Also, the internal boundaries induce a partition of the workspace into regions and each such region is associated with several sets $Ra_{ij}$. For the cuspidal robot at hand, the inner region is associated with the four sets $Ra_{11}$, $Ra_{12}$, $Ra_{21}$ and $Ra_{22}$ (in gray in Figure~\ref{fig16}). The first two, $Ra_{11}$ and $Ra_{12}$, are in $A_1$, while $Ra_{21}$ and $Ra_{22}$ are in $A_2$. The outer region is associated with the two sets $Ra_{13}$ and $Ra_{23}$ that belong to $A_1$ and $A_2$, respectively.
The sets $Ra_{ij}$ are used to determine the maximal uniqueness domains. First, it can be proved that the sets $Ra_{ij}$ are uniqueness domains \citep{Wenger04}. Second, there exist larger uniqueness domains in the joint space. In effect, Figure~\ref{fig16} shows that each aspect is made of three sets $Ra_{ij}$, two of them being associated with the same region in the workspace. If one removes one of these two sets and its boundary from the aspect, the remaining domain is a uniqueness domain \citep{Wenger04}.
Thus, there is still a unique inverse geometric solution in the domain defined by $Qu_1=A_1\dot{-}C(Ra_{12})$ as well as in $Qu_2=A_1\dot{-}C(Ra_{11})$ ($\dot{-}$ means the difference between sets, $C(Ra_{ij})$ means the closure of $Ra_{ij}$). In the same way, $Qu_3=A_2\dot{-}C(Ra_{22})$ and $Qu_4=A_2\dot{-}C(Ra_{21})$ are still uniqueness domains. In addition, these uniqueness domains are maximal \citep{Wenger04}. Figure~\ref{fig17} depicts the four uniqueness domains $Qu_1$, $Qu_2$, $Qu_3$ and $Qu_4$.
		The uniqueness domains can be determined numerically. A simple numerical method can be found in \citep{Wenger04} and will not be reported here.

\begin{figure}[htbp]
            \centering
        \includegraphics[width=0.6\linewidth]{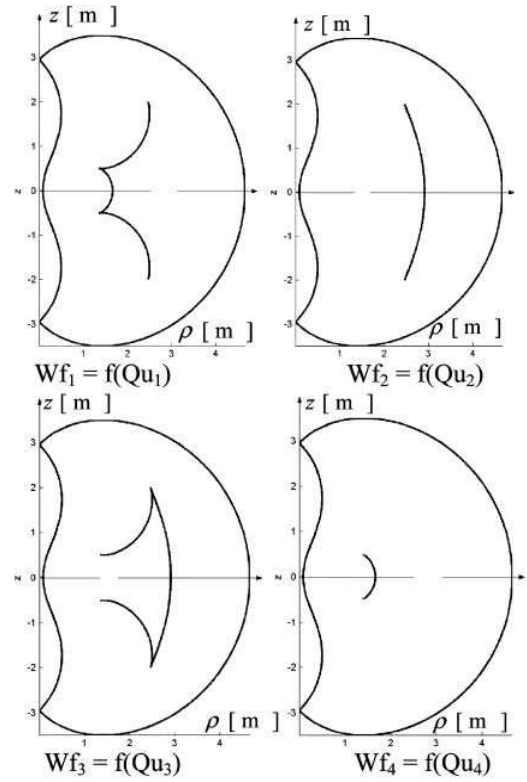}
        \caption{The four maximal regions of feasible paths for the robot of Figure~\ref{fig1}.}
	\label{fig18}
	    \end{figure}

Now that we have determined the (maximal) uniqueness domains, it is then possible to obtain the (maximal) regions of feasible paths in the workspace for a cuspidal robot. These regions are the images in the workspace of the uniqueness domains. Figure~\ref{fig18} shows the four regions $Wf_i=f(Qu_i)$, $i=1,…,4$ of feasible paths for the cuspidal robot at hand. The internal segments that appear in each region indicate sets of points that do not belong to this region and, in turn, segments that the manipulator cannot cross. In fact, each region of feasible paths is the image of an aspect minus one or several segments of the internal boundary surface. This is because the maximal uniqueness domains are defined from the aspects by removing one set $Ra_{ij}$ together with its boundary. By doing so, one removes a part of the boundary surface in the Cartesian space. Note that the four regions $Wf_i$ define regions where any arbitrary path is feasible but they do not feature the full model of feasible paths. In effect, it is always possible to define a feasible path that undergoes a non-singular change of posture in aspect $A_1$ (resp. in aspect $A_2$). In this case, the path would start in $Wf_1$ (resp. in $Wf_3$) and stop in $Wf_2$ (resp. in $Wf_4$). In fact, the full model of feasible paths is obtained when $Wf_1$ and $Wf_2$ (resp. $Wf_3$ and $Wf_4$) are properly “glued” together. To better visualize this fact, the reader is invited to come back to Figure~\ref{fig12}, which precisely shows the full model of feasible paths.

\CCLsubsubsection{Note}: the above definitions and formula stand for any non-redundant robot, with or without joint limits. In \citep{Wenger04}, a cuspidal 3-R robot with joint limits has been analyzed with this theory. It is even possible to include the influence of collisions (self-collisions or collisions with the environment) in the calculus, as reported in \citep{Wenger93}. Practically, however, it is difficult to handle robots with more than 3 joints.
\CCLsection{Identification, enumeration and classification of cuspidal and non-cuspidal robots}
		\CCLsubsection{Simplifying geometric conditions}
		\label{sec:simplify}
		Because of the more complex behavior of a cuspidal robot and because of the difficulty in modeling its kinematic properties, industrial robots should be preferably non-cuspidal. Thus, before designing an innovative kinematic architecture, robot manufacturers should have guidelines and design rules to help them.  Why a manipulator with a given geometry is cuspidal on non-cuspidal has long been a very intriguing question. It is worth noting that this question remains not completely solved. One of the pioneer contributors to this problem was J. Burdick, who observed that under simplifying geometric conditions such as intersecting or parallel joint axes, a 3-R manipulator was non-cuspidal \citep{Burdick95}. Other simplifying conditions were exhibited later \citep{Wenger97} as a direct consequence of the condition recalled in next section. Finally, the following seven geometric conditions were found to define non-cuspidal 3R manipulators:
			
		\begin{enumerate}
\item first two joint axes are parallel; 
\item last two joint axes are parallel; 
\item first two joint axes intersect; 
\item last two joint axes intersect; 
\item first two joint axes are orthogonal and all joint offsets vanish; 
\item the joint axes are mutually orthogonal and the first joint offset vanishes;
\item the joint axes are mutually orthogonal and $d_4^2>d_3^2 ({1+(\frac{r_2}{d_2-d_3})^2)}$. 
\end{enumerate}

These conditions also hold for 6-R manipulators with a spherical wrist because the singularity conditions for the wrist and the regional structure can be decoupled. It is worth noting that conditions 2. and 3. are encountered in most industrial 6-R manipulators. However, the last three conditions are unusual.

A design methodology taking into account the aforementioned geometric conditions was proposed by \citet{Wenger99}.		
		\CCLsubsection{Identification of cuspidal robots and classification}
		\label{sec:identif}
The identification of cuspidal robots is one of the most important issues to solve. When a new robot is designed or even when an existing robot is modified, it is necessary to know if the resulting robot will be cuspidal or not before using it. 
The first important result in the identification of cuspidal robots was found in 1995: the existence of a point in the workspace where three inverse geometric solutions coincide indicates that the robot is cuspidal \citep{ElOmri95}. The formal mathematical proof is not detailed here, it relates on the facts reported in section \ref{sec:non-sing_sing}. A point with three coincident inverse geometric solutions can be identified in a planar cross-section of the workspace as a cusp point (hence the word \text{“}cuspidal\text{”} robot). Moreover, the existence of a point with three coincident solutions can be verified using the univariate polynomial derived to solve the inverse geometric model. For a 3-R manipulator, the existence of cusps can be determined from its fourth-degree inverse kinematic polynomial $P(t)=at^4+bt^3+ct^2+dt+e$ in $t=tan(\frac{\theta_3}{2})$  whose coefficients are function of the geometric parameters and of the variables $R=x^2+y^2$ and $Z=z^2$  (see \citep{Kholi85} for more details on the derivation and properties of this polynomial). The condition for $P(t)$ to have three equal roots can be set as follows:

\begin{equation}\label{eq:syst3}
\left\lbrace\begin{array}{l}
P(t,{d_2},{d_3},{d_4},{\alpha _2},{\alpha _3},{r_2},R,Z) = 0\\
\frac{{\partial P}}{{\partial t}}(t,{d_2},{d_3},{d_4},{\alpha _2},{\alpha _3},{r_2},R,Z) = 0\\
\frac{{{\partial ^2}P}}{{\partial {t^2}}}(t,{d_2},{d_3},{d_4},{\alpha _2},{\alpha _3},{r_2},R,Z) = 0
\end{array} \right.
\end{equation}

where $d_2$, $d_3$, $d_4$, $r_2$, $\alpha_2$, and $\alpha_3$ are the geometric parameters of the robot. If at least one solution to this system exists, this means that the robot is cuspidal. More interestingly, one can try to derive a condition on the geometric parameters for the above system to have (real) solutions. This is very challenging but by doing so, a condition for a 3-R robot to be cuspidal could be found. One needs to eliminate the three variables $t$, $R$ and $Z$. This task is not tractable in the general case but fortunately it becomes feasible (although still complex) once the angle values $\alpha_2$ and $\alpha_3$ are assigned to $90^{\circ}$, i.e. in the particular case of orthogonal manipulators. We can go even further by looking for the conditions under which the number of real solutions changes. By doing so, one will obtain a set of bifurcation surfaces in the parameter space of orthogonal 3-R manipulators where the number of cusp points changes. Accordingly, such bifurcating surfaces can be regarded as sets of transition robots that divide the parameter space into domains where all robots have the same number of cusp points. The algebra involved in system \eqref{eq:syst3} is too complex to be handled by commercial computer algebra tools. \citet{Corvez04} resorted to sophisticated computer algebra tools to solve system \eqref{eq:syst3} by normalizing with $d_2=1$ (without loss of generality) and first considering the more particular case $r_3=0$ (no offset along the last joint axis like the robot shown in fig. 1). They used Groebner Bases and Cylindrical Algebraic Decomposition \citep{Lazard04}, \citep{Collins75} to find the equations of the bifurcating surfaces and the number of domains generated by these surfaces. A kinematic interpretation of this theoretical work was conducted by \citet{Baili03}: the authors analyzed global kinematic properties of one representative manipulator in each domain. Finally, only five different cases were found to exist and the true bifurcating surfaces were shown to take on the following explicit form \citep{Baili03}:

\begin{equation}\label{eq:4}
C_1:~~{d_4} = \sqrt {\frac{1}{2}({d_3}^2 + {r_2}^2 - \frac{{{{({d_3}^2 + {r_2}^2)}^2} - ({d_3}^2 - {r_2}^2)}}{{AB}})} 
\end{equation}

\begin{equation}\label{eq:5}
	C_2:~~{d_4} = \frac{{{d_3}}}{{1 - {d_3}}}B\,\,\,{\text{and}}\,\,\,\,{d_3} < 1
\end{equation}

\begin{equation}\label{eq:6}
C_3:~~{d_4} = \frac{{{d_3}}}{{{d_3} - 1}}B\,\,\,\text{and}\,\,\,\,{d_3} > 1
\end{equation}

\begin{equation}\label{eq:7}
C_4:~~{d_4} = \frac{{{d_3}}}{{1 - {d_3}}}B\,\,\,\text{and}\,\,\,\,{d_3} < 1
\end{equation}

where:
\begin{equation}\label{eq:8}
A = \sqrt {{{({d_3} + 1)}^2} + {r_2}^2} \,\,\,\text{and}\,\,B = \sqrt {{{({d_3} - 1)}^2} + {r_2}^2} 
\end{equation}

\begin{figure}[htbp]
            \centering
        \includegraphics[width=0.8\linewidth]{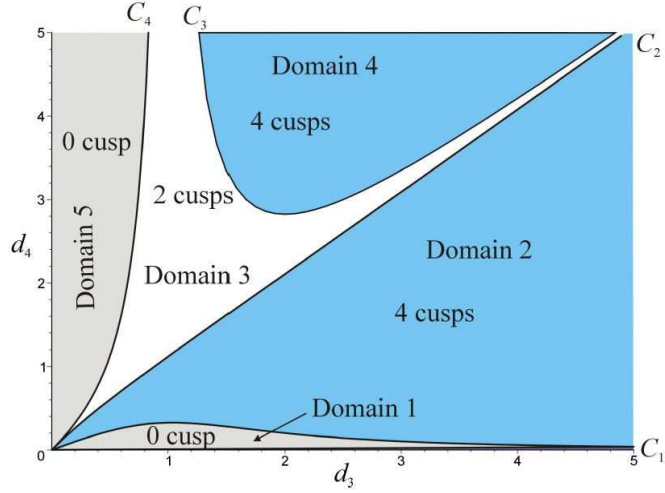}
        \caption{The four bifurcating surfaces and the five domains in the parameter space (section $r_2=1$).}
	\label{fig19}
	    \end{figure}

\begin{figure}[htbp]
            \centering
        \includegraphics[width=1\linewidth]{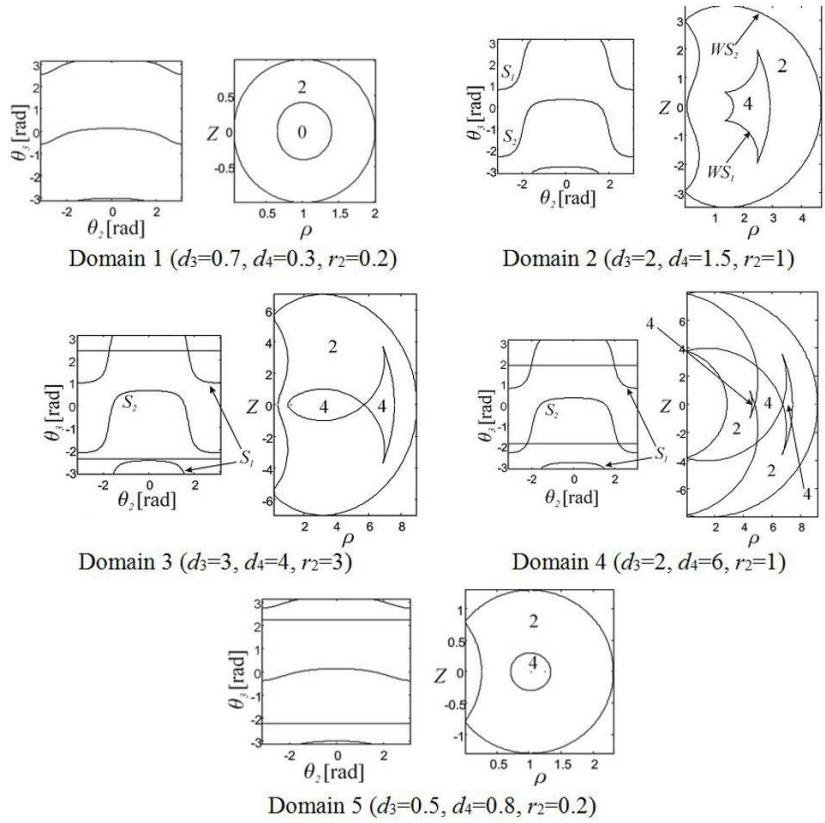}
        \caption{Workspace topologies in each domain.}
	\label{fig20}
	    \end{figure}

These four surfaces divide the parameter space into five domains with 0, 2 or 4 cusps. Figure~\ref{fig19} shows the plots of the surfaces in a section $(d_3, d_4)$ of the parameter space for $r_2=1$. The separating surfaces, which appear as curves there, are labelled with $C_i$ in accordance with the labels in equations \eqref{eq:4} to \eqref{eq:8}. Plotting sections for different values of $r_2$ changes the size of each region but the general pattern does not change and the number of cells remains the same. There are two domains of non-cuspidal manipulators (domains 1 and 5), two domains of cuspidal manipulators with four cusps (domains 2 and 4) and one domain of cuspidal manipulators with two cusps.
			Figure~\ref{fig20} shows the cross sections of the workspace and the singular curves in the joint space, for one representative robot in each domain of the partition. The number of inverse geometric solutions in each region of the workspace is indicated. Figure~\ref{fig20} shows that robots in domain 1 have only two inverse geometric solutions. Also, they have a void in their workspace and they are non-cuspidal. In fact, it can be shown that all other robots have 4 inverse geometric solutions. The other non-cuspidal robots are in domain 5. They have a region with 4 inverse geometric solutions and no void. It is interesting to note that robots in domain 5 were in fact already identified in \citep{Wenger97}, they correspond to the case 7 enumerated in section \ref{sec:simplify}.

The partition of the parameter space and the equations of the bifurcating surfaces allow us to define an explicit necessary and sufficient condition for an orthogonal manipulator with no offset along its last joint axis to be cuspidal. Indeed, Figure~\ref{fig19} shows that a manipulator is cuspidal if and only if it belongs to domains 2, 3 or 4. Thus, an orthogonal manipulator with no offset along its last joint axis is non-cuspidal if and only if:
\begin{equation}\label{eq:syst9}
\left\lbrace\begin{array}{l}
{d_4} > \sqrt {\frac{1}{2}\left( {d_3^2 + r_2^2 - \frac{{{{\left( {d_3^2 + r_2^2} \right)}^2} - d_2^2\left( {d_3^2 - r_2^2} \right)}}{{\sqrt {{{\left( {{d_3} + {d_2}} \right)}^2} + r_2^2} \sqrt {{{\left( {{d_3} - {d_2}} \right)}^2} + r_2^2} }}} \right)}\\
\text{and}\\
{d_3} > {d_2}~\text{or}~\left( {{d_3} < {d_2}~\text{and}~{d_4} < \frac{{{d_3}}}{{{d_2} - {d_3}}} \sqrt {{{\left( {{d_3} - {d_2}} \right)}^2} + r_2^2} } \right)
\end{array} \right.
\end{equation}
It is worthnoting that this necessary and sufficient condition on the geometric parameters is not straighforward and not intuitive at all. 

\begin{figure}[htbp]
            \centering
        \includegraphics[width=0.9\linewidth]{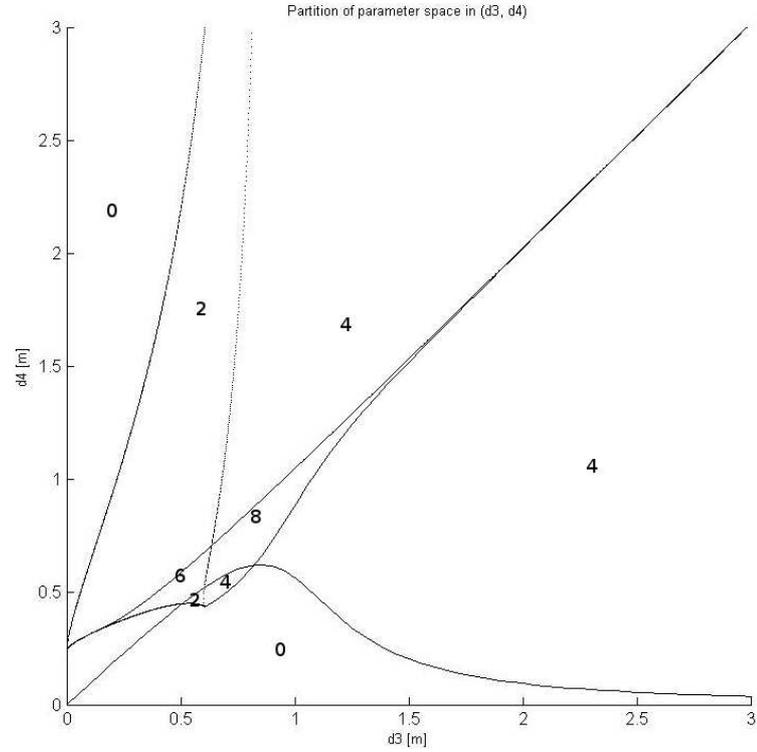}
        \caption{Partition of the parameter space for 3-R robots with joint offsets, section at $r_2=0.3, r_3=0.8$.}
	\label{fig21}
	    \end{figure}

\begin{figure}[htbp]
            \centering
        \includegraphics[width=0.55\linewidth]{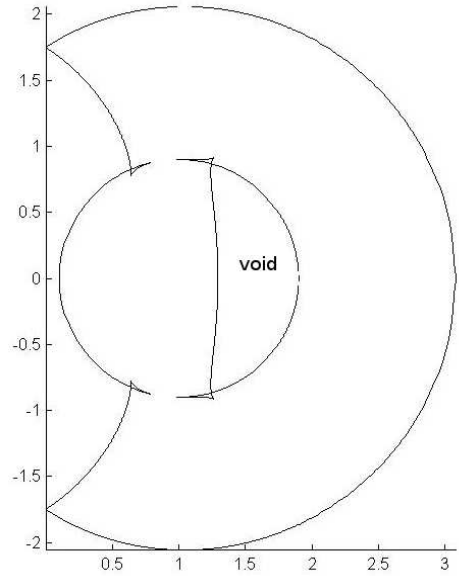}
        \caption{A 3-R orthogonal robot with 8 cusps ($d_2=1$, $d_3=0.91$, $d_4=0.94$, $r_2=0.3$, $r_3=0.9$).}
	\label{fig22}
	    \end{figure}

\CCLsubsubsection{Case $r3\neq0$:} the case of non-zero offset along the last joint axis has also been studied. The number of parameters to handle is 4 instead of 3 and the bifurcating surface equations get much more complicated: the equation of one of the bifurcating surfaces is a $12^{th}$--degree polynomial in the square of the geometric parameters and contains 536 monomials! Because of this complexity, it is difficult to derive an algebraic condition like \eqref{eq:syst9} for a robot to be cuspidal . Figure~\ref{fig21} shows a section of the parameter space at $r_2=0.3$ and $r_3=0.8$. Numerals indicate the numbers of cusps in the domains. It turns out that a robot with non-zero joint offset along its last joint axis may have up to 8 cusps. Such a robot is shown in Figure~\ref{fig22}.
			
\begin{figure}[htbp]
            \centering
        \includegraphics[width=0.5\linewidth]{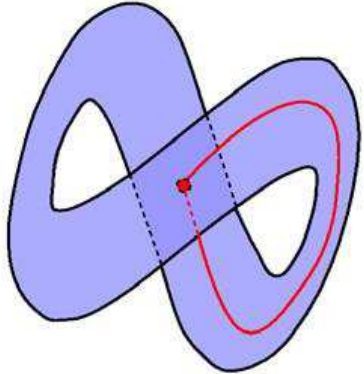}
        \caption{A workspace pattern that would allow a non-singular posture changing path in the absence of cusps.}
	\label{fig23}
	    \end{figure}

			\CCLsubsubsection{Note1}\label{rq:note1}: the existence of a cusp point is a sufficient condition for a robot to be cuspidal but it is not necessary in general. The reason is that the presence of a cusp point provides a way of defining a non-singular change of posture in a local way (the encircling path can be made as close as possible to the cusp point). In theory, there could exist a more global way of defining a non-singular change of posture. Figure~\ref{fig23} shows a workspace pattern that could enable a non-singular change of posture in the absence of any cusp point. This global non-singular change of posture can be interpreted using a comparison with a figure-eight race track where the crossroads is realized with a bridge, i.e. there are two levels at the crossroads. Each level is associated with a posture. Starting from the crossroads on the bridge (level 1), the track is followed until the crossroads at the same horizontal position but under the bridge is reached (level 0).

			In fact, it turns out that the existence of a cusp point is a necessary and sufficient condition for a 3-R orthogonal robot to be cuspial and consequently the pattern shown in Figure~\ref{fig23} cannot exist for these robots. This can be shown by verifying that for one arbitrary robot in each domain of the parameter space associated with no cusp, there is only one inverse geometric solution in each aspect. It is likeable that this result still stands for general 3-R robots (without orthogonal joint axes).
			On the other hand, a planar parallel cuspidal robot was shown in \citep{Coste14} to feature a joint space pattern as in Figure~\ref{fig23} (see section \ref{sec:para}).
			
			\CCLsubsubsection{Note2}: 3-dof robots with one prismatic joint can be classified in a similar way. In fact, the equations will be simpler and such robots are more likely to be non-cuspidal because their inverse geometric model can be often solved in a cascade of two quadratics \citep{Pieper68}. Note that any 3-dof robot with more than one prismatic joint is always non-cuspidal because its inverse geometric model admits two solutions at most (only one for a 3-P robot) \citep{Pieper68}.
			
\begin{figure}[htbp]
            \centering
        \includegraphics[width=0.5\linewidth]{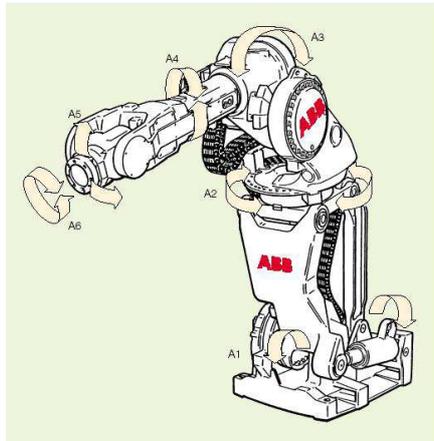}
        \caption{The IRB 6400C robot.}
	\label{fig24}
	    \end{figure}

\begin{figure}[htbp]
            \centering
        \includegraphics[width=0.5\linewidth]{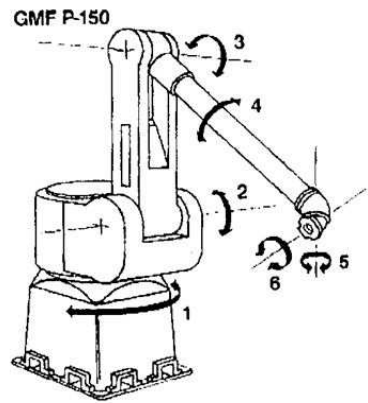}
        \caption{The GMF P150 robot.}
	\label{fig25}
	    \end{figure}

\begin{figure}[htbp]
            \centering
        \includegraphics[width=0.5\linewidth]{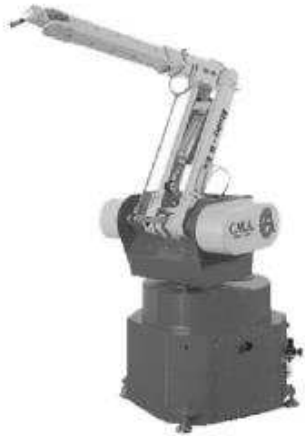}
        \caption{The ROBOX robot.}
	\label{fig26}
	    \end{figure}

			\CCLsection{Higher-degree-of-freedom robots and parallel robots}			
		\CCLsubsection{6-dof serial robots}
		\label{sec:6dof}
		The results pertaining to 3-R robots also hold for 6-R robots with a spherical wrist (i.e, with their last three joint axes intersecting at a common point) because the singularity analysis of the wrist can then be decoupled from that of the regional structure. In section \ref{sec:quest}, we reported the story of the IRB 6400C robot. This robot, shown in Figure~\ref{fig24}, has a spherical wrist and its regional structure is an orthogonal 3R robot that can be shown to be cuspidal.
			Note that the main objective of this new robot design was to save space along the assembly lines and this is why its first joint axis is horizontal instead of vertical \citep{Hemmingson96}. This was a good idea but at the time the engineers of ABB designed their new robot, the classification results were not published. It would be interesting to attempt a new design, keeping the orthogonal architecture with its first axis horizontal but tuning the length parameters in order that the robot falls in one of the interesting classes of non-cuspidal orthogonal robots described by \citet{Zein06a}. 
On the other hand, there is no general result about the enumeration of cuspidal 6-dof manipulators with non-spherical wrist. One of the reasons is the difficulty in analyzing the singularities of general 6-R robots, which depend on four joint variables instead of two in 3-R robots. We think that 6-R manipulators are very likely to be cuspidal, even if the simplifying geometric conditions listed in section \ref{sec:simplify} are satisfied. This is because the inverse kinematics of most 6-dof manipulators with non-spherical wrist is a polynomial of degree higher than 4, which is more likely to admit triple roots. Further research work is required before stating more definitive results but several examples of simple 6-R cuspidal robots with non-spherical wrist exist. One of these robots is the GMF P150 shown in Figure~\ref{fig25} used in the automotive industry for car painting (a similar version exists by COMAU). This robot is close to a PUMA robot, the only difference being the presence of a wrist offset. El Omri showed that without taking account the joint limits, this robot has 16 inverse kinematic solutions and only two aspects \citep{ElOmri96}. Thus, it is cuspidal. Another example is the ROBOX painting robot studied by \citet{Zoppi02} (see Figure~\ref{fig26}). The kinematic architecture is very close to the GFM P150 but the wrist offset is not along the same wrist axis. This cuspidal robot has also 16 inverse kinematic solutions and only two aspects.
			
\begin{figure}[htbp]
            \centering
        \includegraphics[width=0.7\linewidth]{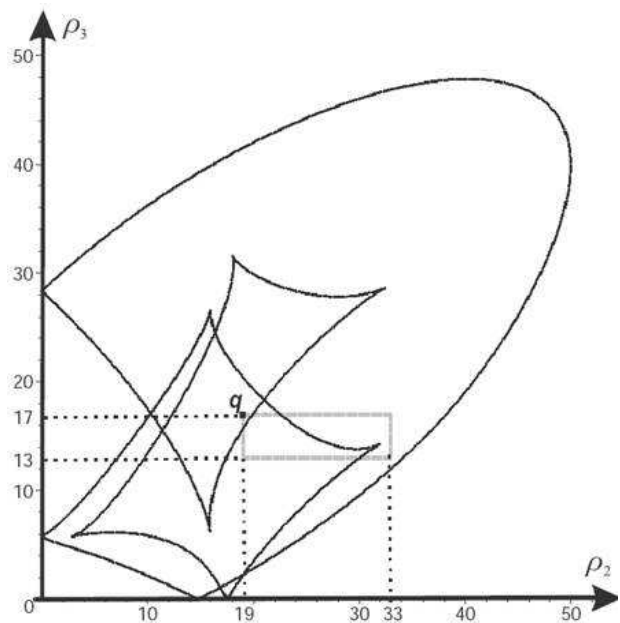}
        \caption{A non-singular assembly-mode changing path encircling a cusp point in a section of the joint space for a 3-R\underline{P}R planar parallel robot.}
	\label{fig27}
	    \end{figure}

If the classification of 6-dof cuspidal and non-cuspidal robots is far from being simple, it is possible to enumerate a set of non-cuspidal robots, namely, those whose inverse geometric polynomial can be solved with quadratics or linear equations in cascade. Such robots were enumerated by \citet{Mavroidis94}.
					\CCLsubsection{Parallel robots}
		\label{sec:para}
		A parallel robot may change its assembly-mode without encountering an output singularity. As first observed in 1998, a parallel robot may be cuspidal in the sense that it may change its assembly-mode (an assembly mode is associated with a solution to the direct geometric problem) without crossing an output singularity \citep{Innocenti98}, \citep{Wenger98}. As shown by \citet{McAree99} and explained in details by \citet{Zein08},  if a parallel robot has 3 coincident assembly modes, thus defining a cusp point in a section of its joint space (not in the workspace this time), then this robot is cuspidal. A non-singular change of assembly-mode can then be accomplished by encircling a cusp point in a section of the joint space \citep{McAree99}, \citep{Zein08}. Figure~\ref{fig27} shows the singularity curves and a non-singular assembly-mode changing path in a section $\rho_1=17$ of the joint space for the 3-R\underline{P}R robot (the underlined letter refers to the actuated joint) analyzed in \citep{Innocenti98} and \citep{McAree99}.

		Because the kinematic equations of a parallel manipulator are very complex, it seems very difficult to derive general geometric conditions for a parallel robot to be cuspidal. However, some results are available for planar parallel robots. One of these results claims that to be cuspidal, a 3-R\underline{P}R robot parallel planar robot should not have similar platform and base triangles \citep{McAree99}, \citep{Kong00} but this is false if the legs are \underline{R}RR instead of R\underline{P}R \citep{Wenger04}. Another interesting result claims that a generic 3-R\underline{P}R robot has always two aspects \citep{Husty09}, \citep{Coste12}. Since such robots have up to six assembly-modes, this shows that they have more than one assembly-mode in one of their aspects, which thus allows them to accomplish non-singular assembly-mode motions. The last interesting result is a proof that a planar parallel robot can be cuspidal while there are no cusp points at all \citep{Coste14}, see \textbf{Note1} in section \ref{sec:identif} and Figure~\ref{fig23}.
		More details on cuspidal planar parallel robots can be found in \citep{Zein06b},  \citep{Zein07} and \citep{Zein08}.
		
		Quite few results exist for 6-dof parallel robots. Out of the octahedral Gough-Stewart platform studied by \citet{McAree99} and J.P. Merlet who showed the existence of non-singular assembly-mode changing motions for this robot, the only other available result, to the author's knowledge, was published in \citep{Caro12}. In this paper, a non-singular assembly-mode changing path was generated for a 6-dof 3-\underline{P}\underline{P}PS parallel robot.
		
		If the interest of designing a cuspidal serial robot is questionable, a parallel robot that is cuspidal is interesting since the robot can thus move in a larger part of its workspace without crossing any output singularity \citep{Wenger04}, \citep{Zein08}. 
		
		\CCLsection{Conclusions}
		The results presented in this chapter can be summarized as follows.
Cuspidal robots, which were first discovered in 1988, have multiple inverse geometric solutions that are not separated by a singular surface.  In the joint space, additional surfaces, called the characteristic surfaces, divide the aspects and separate the solutions. These surfaces are used to define new uniqueness domains and regions of feasible paths in the workspace. The definitions are general and stand for any serial, non-redundant robot with or without joint limits. For 3-dof robots, it is possible to calculate and plot these characteristic surfaces, uniqueness domains and regions of feasible paths. If the first joint is revolute and unlimited, 2-dimensional plots are sufficient. Because there is no simple algebraic definition, these sets must be calculated numerically.

A 3-R robot is non-cuspidal as soon as its first two or last two joint axes are parallel or intersect, or if the three joint axes are mutually orthogonal and the first joint offset is equal to zero. But an orthogonal manipulator with its last joint offset equal to zero may be cuspidal.

A robot is cuspidal if there is at least one point where three inverse geometric solutions coincide. For 3-dof robots, such points appear as cusp points in a cross section of the workspace. In general, the existence of a cusp point is only sufficient for a robot to be cuspidal but for the particular case of 3-R orthogonal robots, this condition is also necessary. The full partition of the parameter space of 3-R orthogonal robots is possible. Bifurcating surfaces divide the parameter space into domains where all robots have the same number of cusp points. 

For 3-R orthogonal robots with no joint offset along their last joint axis, it is possible to derive an explicit DH-parameter based necessary and sufficient condition for a robot to be cuspidal. For general 3-R orthogonal manipulators, the classification is much more complex and does not lend itself to explicit conditions.

Little research work has been conducted on 6-R cuspidal robots. It appears that 6-R robots with non-spherical wrist are very likely to be cuspidal, even if two joint axes intersect or are parallel. However, there is still much work to do before having definitive geometric conditions for general 6-R robots. Resorting to some transversality theorems used by singularity theorists would help going further, providing that we remain in the generic case \citep{Pai91}, \citep{Donelan99}, \citep{Donelan10}. So the first step would be to enumerate all 6-R generic robots.

Finally, the issue of classifying cuspidal and non-cuspidal parallel robots is still much more difficult and remains open. As shown in \citep{Coste14}, a parallel robot may be cuspidal without having any cusp, so that we lack a necessary and sufficient condition for the classification of parallel robots.

\bibliographystyle{plainnat}
\bibliography{CuspidalRobots}   

\begin{thebibliography}{38}
\providecommand{\natexlab}[1]{#1}
\providecommand{\url}[1]{\texttt{#1}}
\expandafter\ifx\csname urlstyle\endcsname\relax
  \providecommand{\doi}[1]{doi: #1}\else
  \providecommand{\doi}{doi: \begingroup \urlstyle{rm}\Url}\fi

\bibitem[Baili et~al.(2003)Baili, Wenger, and Chablat]{Baili03}
M.~Baili, P.~Wenger, and D.~Chablat.
\newblock Classification of one family of 3\uppercase{R} positioning
  manipulators.
\newblock In \emph{Proceedings of 11th Int. Conf. on Advanced Robotics}, 2003.

\bibitem[Borrel and Liegeois(1986)]{Borrel86}
P.~Borrel and A.~Liegeois.
\newblock A study of manipulator inverse geometric solutions with application
  to trajectory planning and workspace determination.
\newblock In \emph{Proceedings of IEEE Int. Conf. Rob. and Aut.}, pages
  1180--1185, 1986.

\bibitem[Burdick(1988)]{BurdickPhD}
J.~W. Burdick.
\newblock \emph{Kinematic analysis and design of redundant manipulators}.
\newblock PhD thesis, Standford University, 1988.

\bibitem[Burdick(1995)]{Burdick95}
J.~W. Burdick.
\newblock A classification of 3\uppercase{R} regional manipulator singularities
  and geometries.
\newblock \emph{Mechanisms and Machine Theory}, 30(1):\penalty0 71--89, 1995.

\bibitem[Caro et~al.(2012)Caro, Wenger, and Chablat]{Caro12}
S.~Caro, P.~Wenger, and D.~Chablat.
\newblock Non-singular assembly mode changing trajectories of a 6-dof parallel
  robot.
\newblock In \emph{Proceedings of ASME Design Engineering Technical Conferences
  and Computers and Information in Engineering Conference}, 2012.

\bibitem[Collins(1975)]{Collins75}
G.E. Collins.
\newblock Quantifier elimination for real closed fields by cylindrical
  algebraic decomposition.
\newblock \emph{Spring lecture, Notes in Computer Science}, 3:\penalty0
  515--532, 1975.

\bibitem[Corvez and Rouillier(2004)]{Corvez04}
S.~Corvez and F.~Rouillier.
\newblock Using computer algebra tools to classify serial manipulators.
\newblock In \emph{Automated Deduction in Geometry, Lectures Notes in Computer
  Science}, pages 31--43, 2004.

\bibitem[Coste(2012)]{Coste12}
M.~Coste.
\newblock A simple proof that generic 3-\uppercase{RPR} manipulators have two
  aspects.
\newblock \emph{ASME J. Mechanisms and Robotics}, 4(1), 2012.

\bibitem[Coste et~al.(2014)Coste, Chablat, and Wenger]{Coste14}
M.~Coste, D.~Chablat, and P.~Wenger.
\newblock Non-singular change of assembly mode without any cusp.
\newblock In J.~Lenar\v{c}i\v{c} and Oussama Khatib, editors, \emph{Advances in
  Robot Kinematics}. Springer, 2014.

\bibitem[Donelan and Gibson(1999)]{Donelan99}
P.S. Donelan and C.G. Gibson.
\newblock Singular phenomena in kinematics.
\newblock In B.~Bruce and D.~Mond, editors, \emph{Singularity Theory}.
  Cambridge University Press, 1999.

\bibitem[Donelan and M\"uller(2010)]{Donelan10}
P.S. Donelan and A.~M\"uller.
\newblock Singularities of regional manipulators revisited.
\newblock In J.~Lenar\v{c}i\v{c} and M.~Stani\v{c}i\'c, editors, \emph{Advances
  in Robot Kinematics, Motion in Man and Machine}. Springer, 2010.

\bibitem[{El Omri}(1996)]{ElOmri96}
J.~{El Omri}.
\newblock \emph{Kinematic Analysis of Robot Manipulators (in French)}.
\newblock PhD thesis, Ecole Centrale de Nantes, 1996.

\bibitem[{El Omri} and Wenger(1995)]{ElOmri95}
J.~{El Omri} and P.~Wenger.
\newblock How to recognize simply a nonsingular posture changing 3-dof
  manipulator.
\newblock In \emph{Proceedings of 7th Int. Conf. on Advanced Robotics}, pages
  215--222, 1995.

\bibitem[Gibson(2000)]{Gibson}
C.G. Gibson.
\newblock Kinematics from the singular viewpoint.
\newblock In J.M. Selig, editor, \emph{Geometrical foundation of Robotics}.
  World Scientific Press, 2000.

\bibitem[Hemmingson et~al.(1996)Hemmingson, Ellqvist, and
  Pauxels]{Hemmingson96}
E.~Hemmingson, S.~Ellqvist, and J.~Pauxels.
\newblock \emph{New Robot Improves Cost-Efficiency of Spot Welding}.
\newblock ABB Review, Spot Welding Robots, 1996.

\bibitem[Husty(2009)]{Husty09}
M.~Husty.
\newblock Non-singular assembly mode change in 3-\uppercase{RPR} parallel
  manipulators.
\newblock In Andreas~M\"uller Andr\'es~Kecskem\'thy, editor, \emph{Singularity
  Theory}. Springer, 2009.

\bibitem[Innocenti and Parenti-Castelli(1998)]{Innocenti98}
C.~Innocenti and V.~Parenti-Castelli.
\newblock Singularity-free evolution from one configuration to another in
  serial and fully-parallel manipulators.
\newblock \emph{ASME J. Mechanical Design}, 120:\penalty0 73--99, 1998.

\bibitem[Khalil and Kleinfinger(1986)]{Symoro}
W.~Khalil and J.~Kleinfinger.
\newblock A new geometric notation for open and closed loop robots.
\newblock In \emph{Proceedings of IEEE Int. Conf. Rob. and Aut.}, pages
  1174–--1179, 1986.

\bibitem[Kholi and Spanos(1985)]{Kholi85}
D.~Kholi and J.~Spanos.
\newblock Workspace analysis of mechanical manipulators using polynomial
  discriminant.
\newblock \emph{ASME J. Mechanisms, Transmission and Automation in Design},
  107:\penalty0 209--215, 1985.

\bibitem[Kong and Gosselin(2000)]{Kong00}
X.~Kong and C.M. Gosselin.
\newblock Determination of the uniqueness domains of 3-\uppercase{RPR} planar
  parallel manipulators with similar platforms.
\newblock In \emph{Proceedings of ASME Design Engineering Technical Conferences
  and Computers and Information in Engineering Conference}, 2000.

\bibitem[Lazard and Rouillier(2004)]{Lazard04}
D.~Lazard and F.~Rouillier.
\newblock \emph{Solving parametric polynomial systems}.
\newblock INRIA Technical Report, 2004.

\bibitem[Mavroidis and Roth(1994)]{Mavroidis94}
C.~Mavroidis and B.~Roth.
\newblock Structural parameters which reduce the number of manipulator
  configurations.
\newblock \emph{ASME J. Mechanical Design}, 116:\penalty0 3--10, 1994.

\bibitem[McAree and Daniel(1999)]{McAree99}
P.R. McAree and R.W. Daniel.
\newblock An explanation of never-special assembly changing motions for 3-3
  parallel manipulators.
\newblock \emph{The International Journal of Robotics Research},
  18(6):\penalty0 556--574, 1999.

\bibitem[Pai and Leu(1991)]{Pai91}
D.K. Pai and M.C. Leu.
\newblock Genericity and singularities of robot manipulators.
\newblock \emph{IEEE Transactions on Robotics and Automation}, 20(4):\penalty0
  545--559, 1991.

\bibitem[Parenti and Innocenti(1988)]{ParentiARK88}
C.V. Parenti and C.~Innocenti.
\newblock Position analysis of robot manipulators : Regions and sub-regions.
\newblock In \emph{Proceedings of Int. Conf. on Advances in Robot Kinematics},
  pages 150--158, 1988.

\bibitem[Pieper(1968)]{Pieper68}
B.~Pieper.
\newblock \emph{The Kinematics of Manipulators Under Computer Control}.
\newblock PhD thesis, Stanford University, 1968.

\bibitem[Wenger(1992)]{Wenger92}
P.~Wenger.
\newblock A new general formalism for the kinematic analysis of all
  non-redundant manipulators.
\newblock In \emph{Proceedings of IEEE Int. Conf. Rob. and Aut.}, pages
  442--447, 1992.

\bibitem[Wenger(1997)]{Wenger97}
P.~Wenger.
\newblock Design of cuspidal and noncuspidal manipulators.
\newblock In \emph{Proceedings of IEEE Int. Conf. Rob. and Aut.}, pages
  2172--2177, 1997.

\bibitem[Wenger(1999)]{Wenger99}
P.~Wenger.
\newblock Some guidelines for the kinematic design of new manipulators.
\newblock \emph{Mechanisms and Machine Theory}, 35(3):\penalty0 437--449, 1999.

\bibitem[Wenger(2004)]{Wenger04}
P.~Wenger.
\newblock Uniqueness domains and regions of feasible paths for cuspidal
  manipulators.
\newblock \emph{IEEE Transactions on Robotics}, 20(4):\penalty0 754--750, 2004.

\bibitem[Wenger and Chablat(1998)]{Wenger98}
P.~Wenger and D.~Chablat.
\newblock Workspace and assembly-modes in fully parallel manipulators: A
  descriptive study.
\newblock In J.~Lenar\v{c}i\v{c} and M.~Husty, editors, \emph{Advances in Robot
  Kinematics}. Kluwer Academic Publisher, 1998.

\bibitem[Wenger and {El Omri}(1993)]{Wenger93}
P.~Wenger and J.~{El Omri}.
\newblock Modeling kinematic properties of type-2 regional manipulators using
  octrees.
\newblock In \emph{Proceedings of IEEE Int. Conf. Man and Cybernetics}, pages
  183--188, 1993.

\bibitem[Whitney(1955)]{Whitney}
H.~Whitney.
\newblock On singularities of mappings of euclidean spaces 1. mappings of the
  plane into the plane.
\newblock \emph{Annals of Mathematics}, 62(3):\penalty0 374--410, 1955.

\bibitem[Zein et~al.(2006{\natexlab{a}})Zein, Wenger, and Chablat]{Zein06a}
M.~Zein, P.~Wenger, and D.~Chablat.
\newblock An exhaustive study of the workspace topologies of all 3\uppercase{R}
  orthogonal manipulators with geometric simplifications.
\newblock \emph{Mechanisms and Machine Theory}, 41(8):\penalty0 971--986,
  2006{\natexlab{a}}.

\bibitem[Zein et~al.(2006{\natexlab{b}})Zein, Wenger, and Chablat]{Zein06b}
M.~Zein, P.~Wenger, and D.~Chablat.
\newblock Singular curves and cusp points in the joint space of
  3-\uppercase{RPR} parallel manipulators.
\newblock In \emph{Proceedings of IEEE Int. Conf. Rob. and Aut.},
  2006{\natexlab{b}}.

\bibitem[Zein et~al.(2007)Zein, Wenger, and Chablat]{Zein07}
M.~Zein, P.~Wenger, and D.~Chablat.
\newblock Singular curves in the joint space and cusp points of
  3-\uppercase{RPR} parallel manipulators.
\newblock \emph{Robotica, special issue on Geometry in Robotics ans Sensing},
  25(6):\penalty0 714--724, 2007.

\bibitem[Zein et~al.(2008)Zein, Wenger, and Chablat]{Zein08}
M.~Zein, P.~Wenger, and D.~Chablat.
\newblock Non-singular assembly-mode changing motions for 3-\uppercase{RPR}
  parallel manipulators.
\newblock \emph{Mechanisms and Machine Theory}, 43(4):\penalty0 480--490, 2008.

\bibitem[Zoppi(2002)]{Zoppi02}
M.~Zoppi.
\newblock Effective backward kinematics for an industrial 6\uppercase{R} robot.
\newblock In \emph{Proceedings of ASME Design Engineering Technical Conferences
  and Computers and Information in Engineering Conference}, 2002.

\end{thebibliography}

\end{document}